\documentclass[letterpaper]{article} 
\usepackage[preprint]{aaai2027}  

\usepackage[hyphens]{url}  
\usepackage{graphicx} 
\usepackage{natbib}  
\usepackage{caption} 
\usepackage{amsmath,amssymb}
\usepackage{algorithm}
\usepackage{algorithmic}

\usepackage{newfloat}
\usepackage{listings}
\DeclareCaptionStyle{ruled}{labelfont=normalfont,labelsep=colon,strut=off} 
\floatstyle{ruled}
\newfloat{listing}{tb}{lst}{}
\floatname{listing}{Listing}

\usepackage{booktabs}
\definecolor{tablegrouptext}{gray}{0.45}

\usepackage{array,multirow,pgfplots}
\pgfplotsset{compat=1.18}

\providecommand{\corresponding}{\thanks{Corresponding author.}}
\title{VideoTok4D: A 4D-Aware Video Tokenizer for Compact World Representation}
\author{Xinyi Chen\equalcontrib, Hanxin Zhu\equalcontrib, Xijun Wang, Xingrui Wang, Sen Liang, Xin Li, Zhibo Chen\corresponding}
\affiliations{University of Science and Technology of China\\
\{chenxinyi0022, hanxinzhu, wangxijun, wxrui\_18264819595, liangsen\}@mail.ustc.edu.cn\\
\{xin.li, chenzhibo\}@ustc.edu.cn}
\date{}

\begin{document}
\maketitle
\pagestyle{plain}
\begin{abstract}
Video tokenizers have emerged as a cornerstone of modern video modeling, underpinning progress in compression, reconstruction and generation by mapping high-dimensional visual signals into compact latent spaces.
However, despite this progress, current tokenization paradigms largely remain within the 2D visual domain, treating videos as image sequences rather than observations of an underlying dynamic 3D world.
Consequently, the learned tokens inherit this observation-centric bias, limiting their capacity to compactly represent real-world 4D scenes.
To mitigate this issue, we propose \textsc{VideoTok4D}, a novel 4D-aware video tokenizer for compact world representation.
Specifically, our approach comprises three key designs: \textbf{1)} a spatiotemporal disentanglement strategy that factorizes videos into static and dynamic tokens for holistic world modeling; \textbf{2)} a track-aware dynamic attention mechanism that aggregates trajectory-aligned cues to promote cross-view motion consistency; and \textbf{3)} \textsc{Co4DGen}, a diffusion prior learned over the resulting \textsc{VideoTok4D} token space for efficient 4D scene generation.
Extensive experiments have demonstrated that our proposed method achieves state-of-the-art performance while requiring up to \textbf{4 orders of magnitude less storage} than dense 4D representations.
Moreover, the compact token space substantially shortens diffusion sequences, enabling efficient generation.

\end{abstract}

\section{Introduction}

As a foundational component of modern video generation, video tokenization transforms high-dimensional frame sequences into compact latent representations, enabling efficient reconstruction and latent generative modeling~\citep{zhao2024cv,tang2024vidtok,mahapatra2025progressive,yang2025cogvideox,kong2024hunyuanvideo,wan2025wan}.

\begin{figure}[t]
    \centering
    \includegraphics[width=\columnwidth]{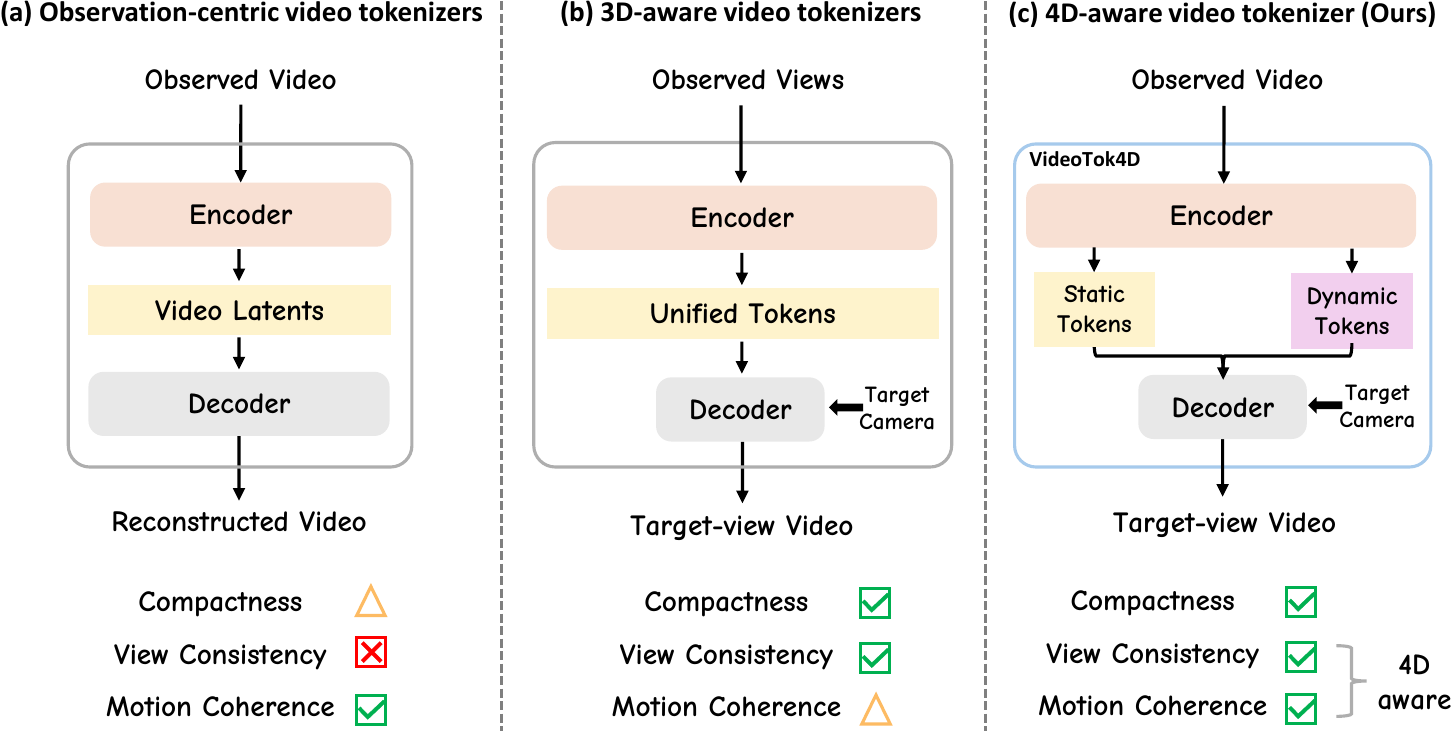}
    \caption{Comparison of video-tokenizer paradigms. (a) Observation-centric tokenizers preserve observed-view motion but lack view consistency. (b) 3D-aware tokenizers are compact and view-consistent but provide limited motion coherence. (c) \textsc{VideoTok4D} achieves all three properties with static and dynamic tokens.}
    \label{fig:representation-comparison}
\end{figure}

Existing video tokenizers can be broadly divided into two paradigms. The first follows a clip-level, observation-centric paradigm, as illustrated in Figure~\ref{fig:representation-comparison}(a), and encodes videos as dense spatiotemporal token grids indexed by image coordinates and time~\citep{zhao2024cv,tang2024vidtok,mahapatra2025progressive}. While effective for high-fidelity reconstruction and compression of observed clips, this design treats videos as sequences of 2D visual measurements rather than projections of an underlying 3D world. Such image-time anchoring yields a view-bound representation, in contrast to a view-invariant scene representation, limiting its ability to maintain coherent geometry and appearance under novel camera trajectories. Moreover, the dense grid structure makes the token budget scale with spatial resolution and temporal duration.
To alleviate this observation-centric limitation, a recent line of 3D-aware tokenizers has adopted a scene-centric paradigm, as illustrated in Figure~\ref{fig:representation-comparison}(b), in which a video is reinterpreted as a collection of posed observations of the underlying 3D scene. Instead of retaining dense image-time tokens, these methods aggregate cross-view information into compact scene representations~\citep{jin2024lvsm,asim2026scenetok}. For example, \textsc{SceneTok} employs novel-view synthesis as supervision, encouraging its compressed tokens to capture geometry- and appearance-consistent 3D scene information.
While these 3D-aware video tokenizers mitigate the view-bound nature of observation-centric tokenizers, they are primarily designed to represent static 3D scenes, where variations across observations are mainly induced by camera motion. In contrast, real-world videos capture dynamic scenes in which the underlying world evolves over time. This temporal evolution introduces an additional dimension that static scene representations do not explicitly model, making it difficult to preserve both view-consistent scene structure and temporally coherent object dynamics beyond the observed viewpoints.

To solve this problem, in this paper we introduce \textsc{VideoTok4D}, a 4D-aware video tokenizer for compact dynamic world representation. Specifically, \textsc{VideoTok4D} consists of three key designs: 1) to compactly represent the dynamic world, a spatiotemporal disentanglement strategy is employed to separate persistent scene content from time-varying object evolution, where the Spatial Branch aggregates stable scene information into static tokens while the Temporal Branch captures temporal object dynamics with dynamic tokens; 2) to preserve coherent object dynamics under changing viewpoints, a track-aware dynamic attention mechanism is employed to isolate independently moving regions after compensating for camera-induced displacement and aggregate their features along estimated trajectories, thereby retaining sparse motion cues and reducing motion blur and ghosting; 3) to validate the learned token space as an efficient representation for conditional 4D world generation, \textsc{Co4DGen}, a diffusion prior over the compact token space, is introduced to sample a joint token state conditioned on a source video and camera trajectories, substantially reducing sampling overhead and enabling multiple synchronized camera trajectories to be rendered from a single sampled state without rerunning the token prior.

Extensive experiments validate \textsc{VideoTok4D} as a compact 4D-aware video tokenizer, showing state-of-the-art dynamic novel-view synthesis with nearly four orders of magnitude less storage than dense 4D representations. \textsc{Co4DGen} further demonstrates the learned token space’s utility for conditional 4D generation, offering a favorable quality--efficiency trade-off.

The main contributions of this paper can be summarized as:
\begin{itemize}
    \item We propose \textsc{VideoTok4D}, a 4D-aware video tokenizer with spatiotemporal disentanglement, enabling compact representation of dynamic worlds beyond the observed viewpoints.
    \item We introduce Track-Aware Dynamic Attention, which models coherent object motion under changing viewpoints by aggregating dynamic features along estimated trajectories.
    \item We develop \textsc{Co4DGen}, a diffusion prior over the learned token space that enables efficient conditional 4D generation by rendering multiple synchronized target-view videos from a single sampled world state.
\end{itemize}

\section{Related Work}

\subsection{Video Tokenization}
Video tokenizers provide the compact latent representations used by modern video generators. Existing approaches reduce spatial and temporal redundancy through image-compatible latent spaces \citep{zhao2024cv}, improved architectures and training \citep{wu2024improved,tang2024vidtok,li2025wf}, or progressive temporal compression \citep{mahapatra2025progressive}. Causal variants are further adopted as the tokenization modules of large-scale video generators \citep{yang2025cogvideox,kong2024hunyuanvideo,hacohen2024ltx,wan2025wan,wang2024omnitokenizer}. Despite favorable compression--reconstruction trade-offs, these tokenizers produce dense spatiotemporal grids optimized for observed clips rather than a scene state reusable across camera trajectories.

Other tokenizers use structured latent layouts to separately encode structure and dynamics or factorize spatial and temporal information \citep{wang2024vidtwin,wang2026vtok}, but remain observation-centric. 3D-aware, scene-centric approaches instead aggregate posed observations: the encoder--decoder variant of \textsc{LVSM} forms fixed implicit scene tokens \citep{jin2024lvsm}, while \textsc{SceneTok} learns compact, diffusable tokens through novel-view supervision \citep{asim2026scenetok}. Both learn representations of static 3D scenes without modeling temporal evolution. \textsc{VideoTok4D} extends this formulation to dynamic scenes, encoding persistent content and object evolution as static and dynamic tokens and aggregating dynamic features along object trajectories.

\subsection{4D Scene Reconstruction}
4D scene reconstruction pursues a related representation goal: capturing persistent scene structure and temporal evolution across viewpoints. Neural fields capture motion through canonical deformation or factorized space--time fields \citep{pumarola2021d,fridovich2023k}, while Gaussian-based methods use time-varying or feed-forward primitives \citep{luiten2023dynamic,yang2024deformable,lin2024gaussian,li2024spacetime,wu20244d,lin2025movies,xu20254dgt,kim2026learning,balice2026no,wang2026flow4dgs}. Other systems regress per-frame pointmaps or dense trajectories \citep{zhang2025monst3r,liu2025trace}, and compression-oriented variants reduce explicit storage \citep{zhang2025mega}. These representations support spatially grounded novel-view rendering, but their states remain distributed over fields, points, Gaussians, or trajectories, with size tied to resolution, duration, or scene complexity. \textsc{VideoTok4D} instead compresses scene content and dynamics into a small latent token set for camera-conditioned reconstruction.

\begin{figure*}[t]
\centering
\includegraphics[width=\textwidth]{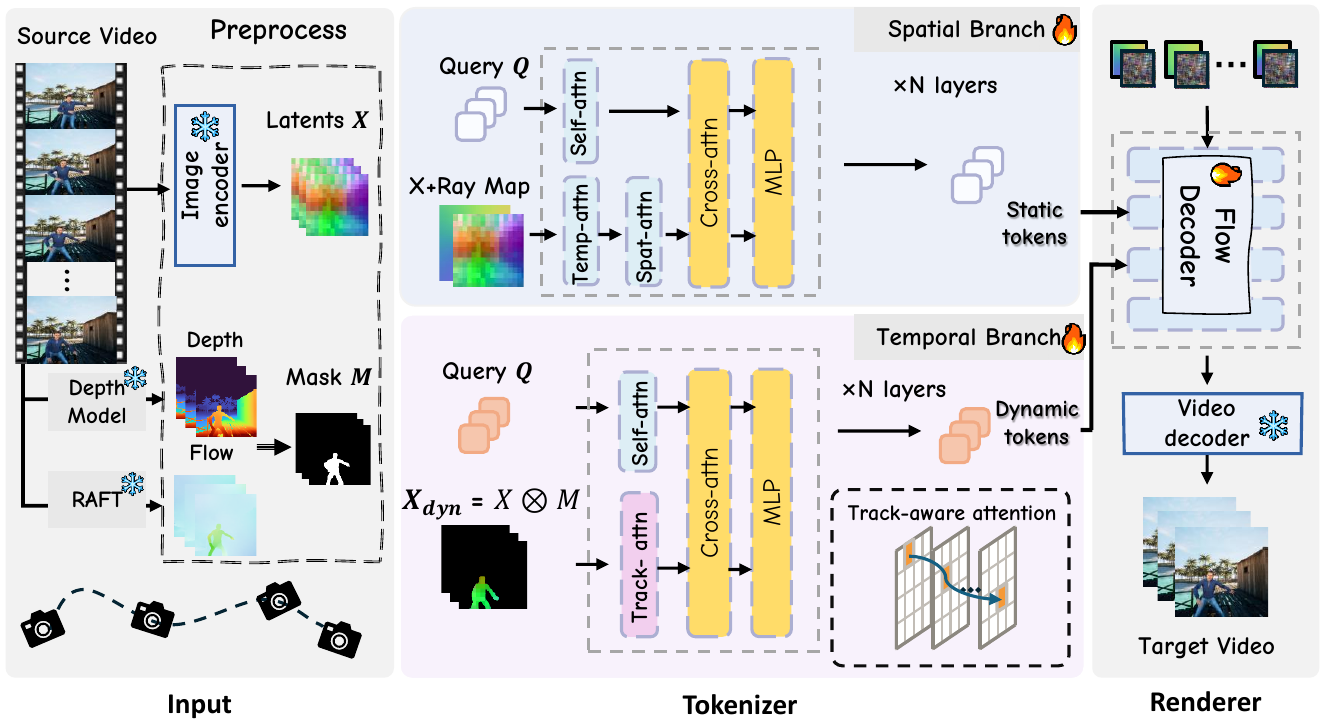}
\caption{Overview of \textsc{VideoTok4D}. From the source video and cameras, the Spatial Branch encodes scene content into static tokens $Z_s$, while the Temporal Branch uses Track-Aware Dynamic Attention to encode object motion into dynamic tokens $Z_d$. The renderer decodes both token groups along a target camera trajectory.}
\label{fig:pipeline}
\end{figure*}

\subsection{4D Scene Generation}

4D scene generation synthesizes dynamic 3D content across viewpoints and time. Explicit-scene methods produce persistent 4D states as neural fields or Gaussian primitives. Optimization-based approaches construct these states from text or images \citep{singer2023text,ren2023dreamgaussian4d,zheng2024unified,bahmani20244d,ling2024align}, while feed-forward methods predict deformable Gaussians from observations \citep{pan2026diff4splat}. Such spatially grounded representations scale with scene detail and duration.

View-space methods generate dense videos along specified camera paths for reconstruction \citep{wu2025cat4d} or direct source-video rerendering \citep{bai2025recammaster,yu2025trajectorycrafter,van2026anyview}. Their view-specific outputs lack a compact scene state reusable across queries. \textsc{Co4DGen} instead samples compact static and dynamic tokens once and reuses the frozen \textsc{VideoTok4D} renderer across camera trajectories.

\section{Method}
\label{sec:method}

\subsection{Overview}

Our goal is to learn a 4D-aware video tokenizer that maps a dynamic video into a compact world representation of the underlying scene rather than the observed views alone. Unlike conventional video tokenizers trained to reconstruct the input clip, \textsc{VideoTok4D} learns its token space through novel-view synthesis, as illustrated in Figure~\ref{fig:pipeline}. Given a source video $V_s=\{I_t^s\}_{t=1}^{T}$, its calibrated camera trajectory $P_s=\{p_t^s\}_{t=1}^{T}$, and a target camera trajectory $P_{\mathrm{tgt}}=\{p_t^{\mathrm{tgt}}\}_{t=1}^{T}$, the encoder produces structured static and dynamic token groups
\begin{equation}
    (Z_s,Z_d)=E_{\phi}(V_s,P_s),
\end{equation}
which are decoded under the target camera trajectory as
\begin{equation}
    \widehat V_{\mathrm{tgt}}
    =D_{\theta}(Z_s,Z_d,P_{\mathrm{tgt}}).
\end{equation}

We further introduce \textsc{Co4DGen}, a diffusion prior that jointly generates static and dynamic tokens. The following sections first describe Spatiotemporal Disentanglement with Track-Aware Dynamic Attention and then introduce conditional generation in the learned token space.

\subsection{Spatiotemporal Disentanglement}

Conventional video tokenizers mix persistent content and time-varying motion in one latent space despite their distinct spatiotemporal support: geometry and appearance span broad regions and frames, whereas object motion is localized and entangled with camera-induced displacement. Dominant persistent evidence can obscure sparse motion cues and hinder coherent trajectory encoding. We therefore encode scene structure and appearance as static tokens and object evolution as dynamic tokens.

Given $(V_s,P_s)$, a frozen VA-VAE image encoder~\citep{yao2025reconstruction} maps each frame $I_t^s$ to a latent feature map $X_t$. The sequence $X_{1:T}=\{X_t\}_{t=1}^{T}$ enters complementary Spatial and Temporal Branches in Figure~\ref{fig:pipeline}. The Spatial Branch augments the full sequence with camera-ray embeddings and applies spatial and temporal attention to construct a camera-aware feature grid. Then, $N_s$ learnable static queries cross-attend to the grid to retrieve global context, forming $Z_s\in\mathbb{R}^{N_s\times d}$. The Temporal Branch applies Track-Aware Dynamic Attention to dynamic-focused features $X_{1:T}^{\mathrm{dyn}}$, with the final states of $N_d$ learnable dynamic queries forming $Z_d\in\mathbb{R}^{N_d\times d}$. The resulting encoding is
\begin{equation}
    \begin{aligned}
        Z_s &= E_s(X_{1:T},P_s),\\
        Z_d &= E_d(X_{1:T}^{\mathrm{dyn}},P_s,F_{1:T-1}),
    \end{aligned}
\end{equation}
where $F_{1:T-1}$ is the inter-frame flow sequence estimated from $V_s$ and used to trace motion across time. The renderer $D_{\theta}$ uses a latent rectified-flow decoder to predict target-view latents conditioned on $(Z_s,Z_d,P_{\mathrm{tgt}})$, which a frozen VideoDCAE decoder~\citep{zheng2025open} maps to RGB frames.

We train the tokenizer with the rectified flow matching objective~\citep{lipman2022flow,liu2023flow} along a linear path from the target-video latent to Gaussian noise. Let $Y^0$ denote the target-video latent in the input space of the frozen VideoDCAE decoder above and $Y^1\sim\mathcal{N}(0,I)$ the noise endpoint. For $\tau\sim\mathcal{U}(0,1)$, we define $Y^\tau=(1-\tau)Y^0+\tau Y^1$ and predict its path velocity as $v_\theta^\tau=v_\theta(Y^\tau,Z_s,Z_d,P_{\mathrm{tgt}},\tau)$. The objective is
\begin{equation}
    \mathcal{L}_{\mathrm{tok}}
    =\mathbb{E}_{\tau,Y^0,Y^1}
    \left\|(Y^1-Y^0)-v_\theta^\tau\right\|_2^2.
\end{equation}

\subsection{Track-Aware Dynamic Attention}

Compressing object dynamics into fixed tokens is difficult for two reasons: pixel displacement mixes camera and object motion, and moving content changes image position, causing fixed-grid temporal aggregation to misalign features. We therefore equip the Temporal Branch with Track-Aware Dynamic Attention. The module proceeds in two stages: pose-compensated motion isolation followed by trajectory-aligned feature aggregation.

\paragraph{Pose-Compensated Motion Isolation.}
Given the source video $V_s$, we use RAFT~\citep{teed2020raft} and UniDepth~\citep{piccinelli2024unidepth} to estimate optical flow $F_{1:T-1}$ and per-frame depth $D_{1:T}$. Together with the relative source-camera transform $T_{t\rightarrow t+1}$, these estimates determine the flow induced by camera motion under a static-scene assumption:
\begin{equation}
    F_t^{\mathrm{pose}}(\mathbf{p})
    =\pi_{t+1}\!\left(
    T_{t\rightarrow t+1}
    \pi_t^{-1}(\mathbf{p},D_t(\mathbf{p}))
    \right)-\mathbf{p}.
\end{equation}
Here $\pi_t^{-1}$ and $\pi_{t+1}$ denote backprojection and projection, respectively. The residual $r_t(\mathbf{p})=\|F_t(\mathbf{p})-F_t^{\mathrm{pose}}(\mathbf{p})\|_2$ isolates object-specific displacement and defines a binary motion mask
\begin{equation}
    M_t(\mathbf p)
    =\mathop{\mathbf{1}}\!\left\{r_t(\mathbf p)>\tau_t\right\},
    \qquad
    X_t^{\mathrm{dyn}}=X_t\odot M_t.
\end{equation}
Here $\mathbf{1}\{\cdot\}$ denotes the indicator function and $\tau_t$ is the residual threshold. The resulting $X_t^{\mathrm{dyn}}$ concentrates the Temporal Branch on independently moving regions while suppressing displacement already explained by the source cameras.

\begin{figure}[t]
\centering
\includegraphics[width=\columnwidth]{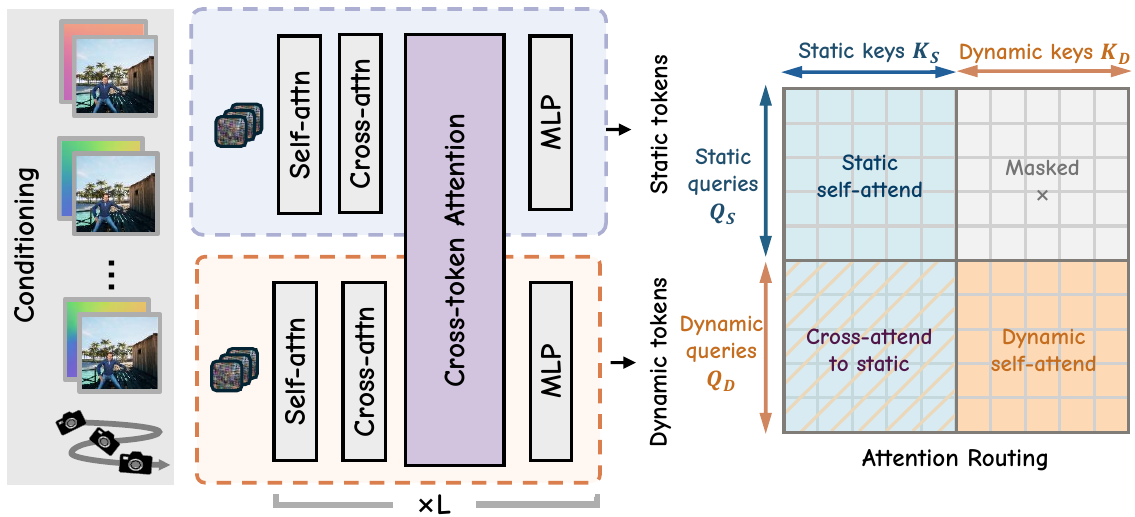}
\caption{Overview of \textsc{Co4DGen}. From a monocular conditioning video and an anchor camera trajectory, the prior jointly denoises static and dynamic tokens with Static--Dynamic Cross-Token Attention.}
\label{fig:co4dgen}
\end{figure}

\paragraph{Track-aware dynamic attention.}
Because moving regions change position, fixed-grid temporal attention can mix unrelated features. We first apply spatial self-attention within each masked dynamic region:
\begin{equation}
    \bar X_t^{\mathrm{dyn}}
    =\mathcal{A}_{\mathrm{sp}}(X_t^{\mathrm{dyn}};M_t),
\end{equation}
where $\mathcal{A}_{\mathrm{sp}}$ denotes spatial self-attention restricted by the motion mask $M_t$. We then sample seeds from the latent-resolution masks and propagate them across frames with the estimated flow. For track $i$, let $\mathcal{V}_i$ denote its valid frame indices. The track update and feature sampling are
\begin{equation}
    p_i^{t+1}=p_i^t+F_t(p_i^t),
    \qquad
    g_i^t=\bar X_t^{\mathrm{dyn}}(p_i^t),
    \quad t\in\mathcal{V}_i.
\end{equation}
Here $g_i^t$ is the feature sampled at the tracked position $p_i^t$, and the resulting trajectory-aligned sequence is $\mathcal{G}_i=\{g_i^t\}_{t\in\mathcal{V}_i}$. Each Temporal Branch layer then applies self-attention independently along every trajectory:
\begin{equation}
    \widetilde g_{i,l}^t
    =\operatorname{MHA}_{l}\!\left(
    g_i^t,\mathcal{G}_i,\mathcal{G}_i
    \right),
    \qquad t\in\mathcal{V}_i,
\end{equation}
where $\operatorname{MHA}_{l}$ denotes self-attention at layer $l$, restricting temporal interaction to features on the same track. Each layer updates the dynamic queries through self-attention, cross-attention to the track features, and an MLP; after $L$ layers, they form $Z_d\in\mathbb{R}^{N_d\times d}$.

\subsection{Co4DGen: Efficient 4D Scene Generation}

\paragraph{Reusable multi-view 4D generation.}
Existing approaches either regenerate dense video latents for individual views or construct dense explicit 4D states for repeated rendering. \textsc{Co4DGen} instead samples one compact 4D token state and reuses the frozen \textsc{VideoTok4D} renderer across synchronized camera trajectories. Token diffusion therefore operates on only $N_s+N_d$ tokens once.

\paragraph{Joint 4D token diffusion.}
Let $\{(V_k,P_k)\}$ denote synchronized videos and calibrated camera trajectories of the same dynamic event. During training, we sample a conditioning camera $c$ and a distinct anchor camera $a$. We define the joint target state as $Z^a=[Z_s^a;Z_d^a]=E_{\phi}(V_a,P_a)$ and learn $p_{\omega}(Z^a\mid V_c,P_c,P_a)$.

\textsc{Co4DGen} treats $Z^a$ as a single diffusion state. At flow time $\tau$, it is perturbed as
\begin{equation}
    Z_{\tau}^a
    =\alpha_{\tau}Z^a+\sigma_{\tau}\epsilon,
    \qquad \epsilon\sim\mathcal{N}(0,I),
\end{equation}
where $\alpha_{\tau}$ and $\sigma_{\tau}$ are schedule coefficients. Conditioning features $h=E_{\mathrm{cond}}(V_c,P_c,P_a)$ are injected into both streams, which interact through Static--Dynamic Cross-Token Attention. At layer $l$, $G_{\omega}^l$ updates the static and dynamic hidden states as $(H_s^{l+1},H_d^{l+1})=G_{\omega}^l(H_s^l,H_d^l,h,\tau)$. Here, $H_s^l$ and $H_d^l$ denote the two hidden streams entering layer $l$, and $G_{\omega}^l$ is the corresponding co-denoising block. Static queries attend only within the static stream, whereas dynamic queries attend to both streams. This asymmetric routing shields static tokens from transient motion while grounding dynamic tokens in scene geometry, appearance, and layout.

\begin{table*}[t]
\centering
\small
\setlength{\tabcolsep}{2.4pt}
\begin{tabular*}{0.98\textwidth}{@{\extracolsep{\fill}}l|cccccccc@{}}
\toprule
\multicolumn{1}{c|}{\raisebox{1.3ex}[0pt][0pt]{Method}} & \shortstack{Repr. Size\\(\#Floats) $\downarrow$} & \raisebox{1.3ex}[0pt][0pt]{PSNR $\uparrow$} & \raisebox{1.3ex}[0pt][0pt]{SSIM $\uparrow$} & \raisebox{1.3ex}[0pt][0pt]{LPIPS $\downarrow$} & \raisebox{1.3ex}[0pt][0pt]{rFID $\downarrow$} & \raisebox{1.3ex}[0pt][0pt]{rFVD $\downarrow$} & \shortstack{Motion\\Smooth. $\uparrow$} & \shortstack{Subject\\Cons. $\uparrow$} \\
\midrule
\multicolumn{9}{@{}l}{\textcolor{tablegrouptext}{\textit{Explicit Representation}}} \\
MoVieS & 948.93M & 19.06 & 0.66 & 0.25 & 34.61 & 492.11 & 0.73 & 0.71 \\
4DGT & 6.72 M & 18.76 & 0.64 & 0.33 & 27.33 & 319.28 & 0.81 & 0.77 \\
C4G & 0.92 M & 14.27 & 0.45 & 0.46 & 31.01 & 378.06 & 0.79 & 0.74 \\
\midrule
\multicolumn{9}{@{}l}{\textcolor{tablegrouptext}{\textit{Latent Representation}}} \\
\textsc{SceneTok}\textsuperscript{$\dagger$} & \textbf{98.30 K} & 17.67 & 0.50 & 0.32 & 22.95 & 247.41 & 0.80 & 0.76 \\
\textsc{Ours (VideoTok4D)} & \textbf{98.30 K} & \textbf{20.22} & \textbf{0.68} & \textbf{0.23} & \textbf{19.65} & \textbf{185.57} & \textbf{0.88} & \textbf{0.81} \\
\bottomrule
\end{tabular*}
\caption{Dynamic novel-view synthesis; Motion Smoothness and Subject Consistency are normalized to $[0,1]$.}
\label{tab:dynamic-nvs}
\end{table*}

The prior predicts the velocity of the joint token state with the objective
\begin{equation}
    \mathcal{L}_{\mathrm{Co4DGen}}
    =\mathbb{E}_{\tau,Z^a,\epsilon}
    \left\|
    \widehat v_{\omega}(Z_{\tau}^a,h,\tau)-v_{\tau}
    \right\|_2^2,
\end{equation}
where $\widehat v_{\omega}$ and $v_{\tau}$ denote the predicted and target joint velocities, respectively. At inference, we draw one joint token state from the conditional prior and reuse it with the frozen renderer $D_{\theta}$ across all query trajectories $P_q$, $q\in\mathcal{Q}$.

\section{Experiments}
\label{sec:experiments}

We evaluate \textsc{VideoTok4D} along three axes: (1) reconstruction fidelity and storage for dynamic novel-view synthesis, (2) generation quality and efficiency through \textsc{Co4DGen}, and (3) the correspondence biases of static and dynamic tokens toward scene content and motion.

\begin{figure*}[!t]
\centering
\includegraphics[width=0.95\textwidth]{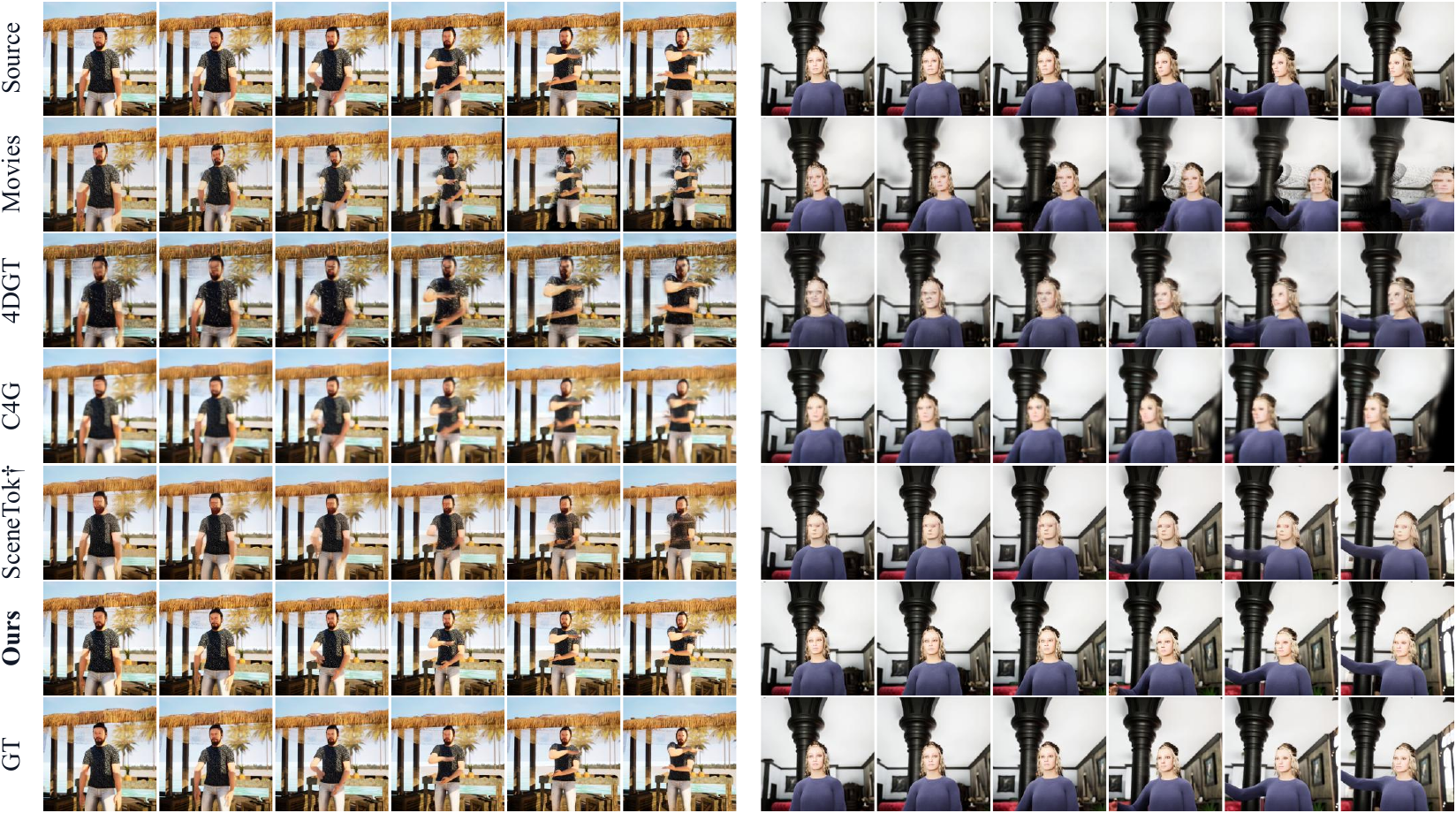}
\caption{Qualitative comparison of dynamic novel-view synthesis across timestamps. \textsc{VideoTok4D} better preserves geometry, appearance, and motion than competing methods.}
\label{fig:dynamic-nvs-qual}
\end{figure*}

\begin{table*}[t]
\centering
\small
\setlength{\tabcolsep}{2.2pt}
\begin{tabular*}{0.98\textwidth}{@{\extracolsep{\fill}}l|cccccccc@{}}
\toprule
\multicolumn{1}{c|}{\raisebox{1.3ex}[0pt][0pt]{Method}} & \shortstack{Seq.\\Length $\downarrow$} & \shortstack{Sampling\\Runs $\downarrow$} & \raisebox{1.3ex}[0pt][0pt]{Time (s) $\downarrow$} & \raisebox{1.3ex}[0pt][0pt]{gFID $\downarrow$} & \raisebox{1.3ex}[0pt][0pt]{gFVD $\downarrow$} & \shortstack{Motion\\Smooth. $\uparrow$} & \raisebox{1.3ex}[0pt][0pt]{FVD-V $\downarrow$} & \raisebox{1.3ex}[0pt][0pt]{CLIP-V $\uparrow$} \\
\midrule
\textsc{DFoT}\textsuperscript{$\ddagger$} & 8,192 & 9 & 405.60 & 48.13 & 342.86 & 0.72 & 334.51 & 0.74 \\
\textsc{SceneGen}\textsuperscript{$\dagger$} & \textbf{1,536} & \textbf{1} & 124.24 & 32.47 & 296.69 & 0.77 & 284.86 & 0.81 \\
\textsc{Ours (Co4DGen)} & \textbf{1,536} & \textbf{1} & \textbf{119.18} & \textbf{25.56} & \textbf{224.94} & \textbf{0.85} & \textbf{210.36} & \textbf{0.84} \\
\bottomrule
\end{tabular*}
\caption{Conditional 4D scene generation from one 32-frame video. Time covers all nine views; sequence length is per run, and Sampling Runs counts independent trajectories. Motion Smoothness and CLIP-V are normalized to $[0,1]$.}
\label{tab:token-generation}
\end{table*}

\begin{figure*}[t]
\centering
\includegraphics[width=\textwidth]{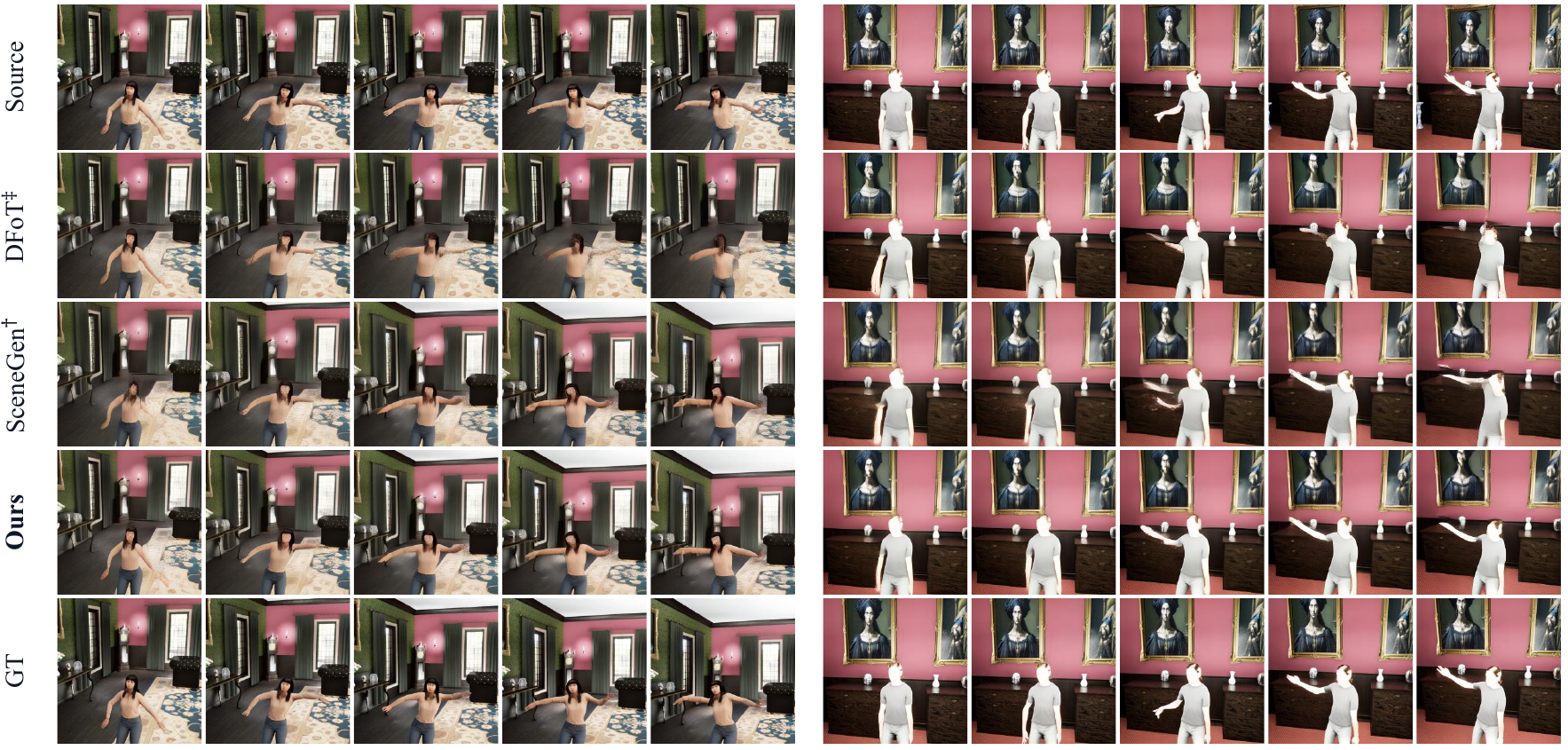}
\caption{Qualitative conditional 4D scene generation. Token-space results show one of nine trajectory-specific decoder outputs obtained from the same sampled 4D state.}
\label{fig:token-generation}
\end{figure*}

\subsection{Experimental Setup}

\paragraph{Dataset.}
We evaluate all methods on the MultiCamVideo-Dataset~\citep{bai2025recammaster}, which contains 13,600 dynamic scenes, each captured by 10 synchronized cameras with calibrated trajectories. We use an asset-disjoint split, holding out 400 dynamic scenes for testing such that no underlying scene assets overlap with the remaining training scenes.

\paragraph{Implementation details.}
We train \textsc{VideoTok4D} on 32-frame clips at $256\times256$ resolution, using $N_s=1024$ static and $N_d=512$ dynamic tokens with dimension $d=64$. Training runs for 100K iterations on 8 NVIDIA A100 GPUs with an effective batch size of 128. We train \textsc{Co4DGen} on 32-frame clips at the same resolution for 300K iterations on 16 NVIDIA A100 GPUs with an effective batch size of 96.

\paragraph{Evaluation protocol.}

All methods use the same held-out scenes, 32-frame source videos, target trajectories, and $256\times256$ resolution. We focus on baselines trainable or adaptable on MultiCamVideo under comparable conditioning and model capacity, treating systems built on large pretrained video priors and external data~\citep{bai2025recammaster,yu2025trajectorycrafter,van2026anyview} as complementary rather than controlled baselines. The \textsuperscript{$\dagger$} mark denotes capacity-matched \textsc{SceneTok}/\textsc{SceneGen} retrained on MultiCamVideo, and \textsuperscript{$\ddagger$} denotes \textsc{DFoT} adapted and retrained under the same protocol. All conditional-generation claims are restricted to this setting.

\subsection{Dynamic Novel-View Synthesis}

Each method encodes the dynamic scene once from a source video and renders a synchronized sequence along a target camera trajectory. Table~\ref{tab:dynamic-nvs} reports PSNR, SSIM~\citep{wang2004image}, LPIPS~\citep{zhang2018unreasonable}, reconstruction FID/FVD (rFID/rFVD)~\citep{heusel2017gans,unterthiner2018towards}, VBench Motion Smoothness and DINO-based Subject Consistency~\citep{huang2024vbench,caron2021emerging}, and per-sequence representation size. Representation size counts the FP32 per-sequence state for 32 frames, excluding shared model parameters; the time-conditioned Gaussian states of MoVieS are materialized at all evaluated timestamps.

\textsc{VideoTok4D} achieves the strongest reconstruction and temporal quality while maintaining a compact scene state. It uses nearly four orders of magnitude less storage than MoVieS and outperforms the capacity-matched \textsc{SceneTok}\textsuperscript{$\dagger$}, suggesting that the structured static--dynamic tokenization is more effective than a unified token set at the same capacity. Figure~\ref{fig:dynamic-nvs-qual} shows the same trend: explicit methods exhibit ghosting, unstable geometry, or oversmoothed motion, while \textsc{SceneTok}\textsuperscript{$\dagger$} produces inconsistent limb motion. In contrast, \textsc{VideoTok4D} preserves the static background while reconstructing coherent arm trajectories with clear boundaries and stable appearance.

\subsection{4D Scene Generation}

We compare \textsc{Co4DGen} with the dense video-diffusion baseline \textsc{DFoT}\textsuperscript{$\ddagger$}~\citep{song2025history} and the capacity-matched token baseline \textsc{SceneGen}\textsuperscript{$\dagger$}~\citep{asim2026scenetok}. Given $(V_c,P_c,P_a)$, \textsc{Co4DGen} samples one 1,536-token 4D world representation and queries the frozen renderer for each of the nine unobserved camera trajectories, producing temporally aligned target-view videos without target-view RGB observations. \textsc{SceneGen}\textsuperscript{$\dagger$} follows the same one-sample, nine-query protocol with unified tokens, whereas \textsc{DFoT}\textsuperscript{$\ddagger$} requires nine independent dense-video diffusion runs.

Table~\ref{tab:token-generation} shows that \textsc{Co4DGen} achieves the best generation FID/FVD (gFID/gFVD), Motion Smoothness, and end-to-end time. FVD-V applies FVD across temporally aligned views, while CLIP-V measures source--target similarity at matched timestamps using CLIP features~\citep{radford2021learning}. \textsc{Co4DGen} is $3.40\times$ faster than \textsc{DFoT}\textsuperscript{$\ddagger$} and outperforms the capacity-matched \textsc{SceneGen}\textsuperscript{$\dagger$}, yielding a better quality--efficiency trade-off. Figure~\ref{fig:token-generation} further shows that \textsc{Co4DGen} preserves scene structure, object appearance, and articulated motion more consistently than the blurred \textsc{DFoT}\textsuperscript{$\ddagger$} outputs and the less stable motion of \textsc{SceneGen}\textsuperscript{$\dagger$}.

\subsection{Analysis of Static--Dynamic Tokens}

To characterize the roles that emerge in the static and dynamic token groups, we examine both the correspondence cues they retain and their influence on reconstruction. We use two probes: frozen-token track decoding measures the static and dynamic correspondence information accessible from each group, while group-wise masking measures how removing either group affects reconstruction fidelity and temporal coherence.

\paragraph{Track decoding from frozen tokens.}
We keep the tokenizer fixed and train identical lightweight track decoders on $Z_s$, $Z_d$, or their combination, using pseudo tracks from \textsc{Trace Anything}~\citep{liu2025trace} as supervision. As shown in Table~\ref{tab:track-decoding}, $Z_s$ produces lower EPE on static tracks, whereas $Z_d$ is more accurate on dynamic tracks. Their combination achieves the lowest EPE across the static, dynamic, and mixed subsets. The largest gain occurs on the mixed subset, where combining $Z_s$ and $Z_d$ reduces EPE from 0.445 for the best individual token group to 0.240. These results reveal distinct but non-exclusive correspondence biases: $Z_s$ favors static structure, $Z_d$ favors object motion, and joint decoding provides more complete scene-wide correspondence cues.

\begin{table}[t]
\centering
\small
\setlength{\tabcolsep}{5.0pt}
\begin{tabular}{lccc}
\toprule
Token Input & EPE-sta $\downarrow$ & EPE-dyn $\downarrow$ & EPE-mix $\downarrow$ \\
\midrule
$Z_s$ & 0.293 & 0.714 & 0.445 \\
$Z_d$ & 0.702 & 0.286 & 0.552 \\
$Z_s+Z_d$ & \textbf{0.221} & \textbf{0.259} & \textbf{0.240} \\
\bottomrule
\end{tabular}
\caption{Track decoding from frozen tokens. EPE-sta and EPE-dyn denote errors on tracks from static and dynamic regions, respectively, while EPE-mix is computed over a mixed track set containing both.}
\label{tab:track-decoding}
\end{table}

\paragraph{Token masking.}
Token masking reveals asymmetric but complementary roles (Table~\ref{tab:token-masking}). Removing $Z_s$ drops PSNR by over 10\,dB and nearly halves Motion Smoothness, showing that it provides the global appearance and geometric scaffold required for stable rendering across views and time. Removing $Z_d$ causes smaller but consistent drops across all metrics, indicating that it refines this scaffold with time-varying motion cues. Together with track decoding, these results support distinct but complementary specialization rather than a strict decomposition; both groups jointly preserve scene structure, motion fidelity, and stable rendering across views and time, demonstrating that neither group alone provides a complete dynamic-scene representation.

\begin{table}[t]
\centering
\footnotesize
\setlength{\tabcolsep}{2.0pt}
\begin{tabular*}{\columnwidth}{@{\extracolsep{\fill}}lcccc}
\toprule
Variant & PSNR $\uparrow$ & SSIM $\uparrow$ & LPIPS $\downarrow$ & Motion Smooth. $\uparrow$ \\
\midrule
w/o $Z_s$ & 10.01 & 0.20 & 0.75 & 0.45 \\
w/o $Z_d$ & 17.23 & 0.46 & 0.35 & 0.79 \\
Full & \textbf{20.22} & \textbf{0.68} & \textbf{0.23} & \textbf{0.88} \\
\bottomrule
\end{tabular*}
\caption{Token masking; Motion Smoothness is in $[0,1]$.}
\label{tab:token-masking}
\end{table}

\subsection{Ablation Studies}

\begin{table}[t]
\centering
\footnotesize
\setlength{\tabcolsep}{1.2pt}
\begin{tabular*}{\columnwidth}{@{\extracolsep{\fill}}lcccc}
\toprule
\raisebox{0.7ex}[0pt][0pt]{Variant} & \raisebox{0.7ex}[0pt][0pt]{PSNR $\uparrow$} & \raisebox{0.7ex}[0pt][0pt]{SSIM $\uparrow$} & \raisebox{0.7ex}[0pt][0pt]{LPIPS $\downarrow$} & \shortstack[c]{Motion\\Smooth. $\uparrow$} \\
\midrule
w/o Disentanglement & 17.78 & 0.53 & 0.31 & 0.79 \\
w/o Motion Isolation & 18.82 & 0.61 & 0.27 & 0.83 \\
w/o Track-Aware Attn. & 18.11 & 0.59 & 0.29 & 0.82 \\
\textsc{VideoTok4D} & \textbf{20.22} & \textbf{0.68} & \textbf{0.23} & \textbf{0.88} \\
\bottomrule
\end{tabular*}
\caption{Ablation of \textsc{VideoTok4D} under the shared protocol and 1,536-token budget; Motion Smoothness is normalized to $[0,1]$.}
\label{tab:tokenizer-ablation}
\end{table}

We retrain three tokenizer ablations under the same protocol. ``w/o Disentanglement'' merges static and dynamic tokens into 1,536 capacity-matched unified tokens; ``w/o Motion Isolation'' removes pose-compensated masking but retains flow-guided tracks; and ``w/o Track-Aware Attention'' replaces trajectory-aligned aggregation with fixed-grid temporal attention. All variants share the token budget, decoder, and training schedule, isolating representation factorization, motion selection, and temporal aggregation.

Table~\ref{tab:tokenizer-ablation} shows that all three designs contribute. Disentanglement has the largest effect: replacing the two token groups with unified tokens reduces PSNR by 2.44\,dB and weakens reconstruction fidelity and temporal coherence, supporting explicit representation factorization. The motion-specific variants further clarify the roles of evidence selection and temporal aggregation. Removing motion isolation causes a consistent degradation, while fixed-grid attention produces a larger drop, suggesting that pose-compensated masking suppresses camera-induced displacement and trajectory-aligned aggregation is important for coherent motion encoding.

\FloatBarrier

\section{Conclusion}

We introduced \textsc{VideoTok4D}, a 4D-aware video tokenizer for compact representation of dynamic worlds beyond the observed viewpoints. At its core is a spatiotemporal disentanglement strategy that represents persistent scene content with static tokens and time-varying object evolution with dynamic tokens. Track-Aware Dynamic Attention preserves coherent object motion under changing viewpoints by compensating for camera-induced displacement and aggregating dynamic features along estimated trajectories. Building on this token space, \textsc{Co4DGen} enables efficient conditional 4D generation by sampling a joint world state once and rendering multiple synchronized target-view videos without rerunning the token prior. Extensive experiments demonstrate state-of-the-art dynamic novel-view synthesis with up to four orders of magnitude less storage than dense 4D representations, together with a favorable quality--efficiency trade-off for conditional 4D generation.

\bibliography{aaai2027}

\end{document}